# Deep CNNs for large scale species classification

Raj Prateek Kosaraju

## Abstract

*Large Scale image classification is a challenging problem within the field of computer vision. As the real world contains billions of different objects, understanding the performance of popular techniques and models is vital in order to apply them to real world tasks. In this paper, we evaluate techniques and popular CNN based deep learning architectures to perform large scale species classification on the dataset from iNaturalist 2019 Challenge. Methods utilizing dataset pruning and transfer learning are shown to outperform models trained without either of the two techniques. The ResNext based classifier outperforms other model architectures over 10 epochs and achieves a top-one validation error of 0.68 when classifying amongst the 1,010 species.*

1. Introduction

The world contains hundreds of thousands of species of plants and animals. Thanks to the prevalence of smart phones, people take photos of plants and animals in the wild but are unable to identify the exact species they belong to without expert knowledge. Identification of the species an animal belongs to can help with understanding them and appreciating them further. With the recent advances in GPUs, deep learning, and availability of large datasets, we can attempt to solve this problem in a computationally feasible way. We will explore the performance of a baseline model, then review past literature and utilize popular CNN architectures as well as methods like transfer learning and data pruning to evaluate classification performance and discover what works best.

2. Related Work

Macoadha proposed an Inception v3 model [9] to tackle the previous year's iNaturalist 2018 competition, achieving a low validation error of 0.40 when trained over 75 epochs. Their work utilizes a pre-trained InceptionNet and finetunes the hyper-parameters carefully to achieve this error.

In other work, Deng et al. [1] investigate issues with large scale image classification and propose interesting findings. They highlight that computations issues become crucial in algorithm design and that convention image classification wisdom does not hold when number of classes increases drastically.

3. Data

3.1. Dataset

The iNat Challenge 2019 contains images of 1,010 species, with a combined training and validation set of 268,243 images. The test set consists of 35,350 images. These images have been carefully collected and verified by multiple members of iNaturalist, an online network of biologists. The competition has been running for previous years, but now features a smaller number of highly similar categories captured in a variety of situations. The dataset is well categorized into Amphibians, Birds, Fungi, Insects, Plants and Reptiles, with a total of 1010 species. The train set, validation set and the test set contain images from all 6 species categories. Each image is of varying shapes, but typically around the size of 800px x 800px.

The number of images per class also varies and is summarized in Table 1. A histogram of this distribution is presented in Figure 1.

| Maximum | Minimum | Median | Standard Dev. |
|---|---|---|---|
| 500 | 212 | 262 | 167 |

Table 1: Statistics on number of images per class

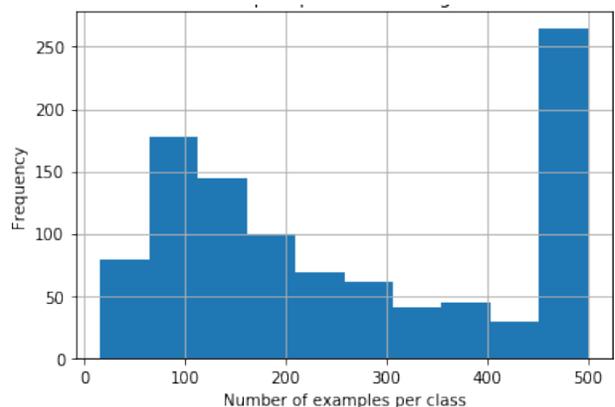

Figure 1: Distribution of number of images per class

### 3.2. Preprocessing

The dataset was preprocessed to resize images such that they are of the appropriate size for the CCN architecture we use. All the architectures we utilized in this paper expect 224 x 224 images and images were resized to this size. We split the train/validation images set into 90% train and 10% validation, yielding 241,418 images for the train set and 26,824 images for the validation set.

As an additional preprocessing step, discarding images from classes which have fewer than 100 images from class was experimented with and the results from this will be described in more detail in a later section.

## 4. Method

### 4.1. Inputs

The input features to the classification model are the images after applying the pre-processing steps described above. The input also includes the integer label that each of these images should be classified as.

### 4.2. Output

The output of the classification model would be the top label labels that the classification model classifies as image as belonging to, sorted by confidence.

### 4.3. Loss Formulation

Since this is a classification task, we chose to optimize on Cross Entropy Loss in all our experiments. While Cross Entropy Loss penalizes especially penalizes predictions where the models was confident but incorrect.

$$Loss = -\sum y_{o,c} \log(p_{o,c})$$

Here, y is the true label and p is the predicted label obtained as an output from the model.

### 4.4. Training Approach

#### 4.4.1 Data Pruning

Based on Figure 1, it's evident that there are many classes with very few images in the dataset. These classes would be difficult for the classifier to perform well on due to the lack of images to learn from. As the validation set has the same distribution of images per class as the train set, we experiment with pruning images from the train set which have fewer than 100 images per class. There are 208 classes with fewer than 100 images per class in the dataset, and this would result in discarding 14,621 images of the 241,418 train set images.

#### 4.4.2 Transfer learning for weights initialization

Transfer Learning has been a popular technique used to improve the performance of classifiers over image classification tasks. We experiment with training models that were pre-trained over the ImageNet 1k dataset and check what the impact is by training all layers and freezing all but the last layer when compared to training the models from scratch.

#### 4.4.3 CNN model architectures

There have been several popular CNN architectures that have been known to perform well on image classification tasks. We test several of them to find which architecture suits large scale image classification the best.

**AlexNet**[6] was the first widely popular CNN architecture that beat traditional image classification techniques. It is composed of five convolutional layers followed by three fully connected layers. It used the ReLU activation function instead of the then standard tanh and sigmoid.

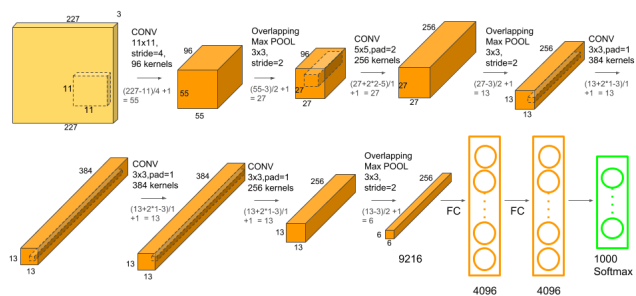

Figure 2 AlexNet architecture adapted from [8]

**Densenet**[4] was built on top of networks proposed earlier to address the vanishing gradients problem. Densenet simply connected every layer to every other layer. This strengthened feature propagation and substantially reduced the number of parameters allowing for deeper networks.

**SqueezeNet**[10] is a smart architecture that provides accuracy similar to AlexNet while being 3 times faster and 50x fewer parameters. One of the ways it achieves this is by using 1x1 convolutions instead of the traditional 3x3 convolutions.

**ShuffleNet**[7] utilizes pointwise group convolution and channel shuffle to reduce computation cost while maintaining accuracy. It beats MobileNet on the ImageNet classification challenge and is 13x faster than AlexNet while maintaining comparable accuracy.

**Resnext**[5] is an extension of the original ResNet deep residual network architecture. It replaces the standard residual block with another one that uses split-transform-merge strategy used in inception modules.

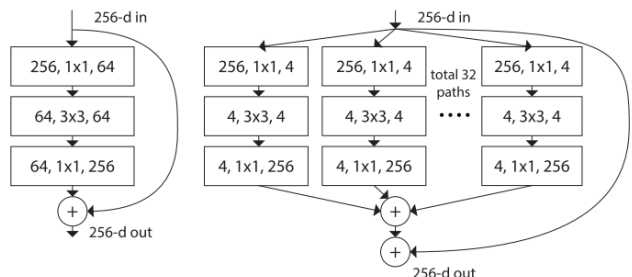

Figure 3: A block from ResNet (left) compared to a block from ResNext (right) from [5]

5. Experiments

This section discusses the results of the various approaches discussed above.
If the predicted label matches the ground truth label then the error for that image is 0, otherwise it is 1. The final validation error rate is the error averaged across all images in the validation set. All models were trained only over 10 epochs to minimize the computational resources required for these experiments.

5.1. Data preprocessing

Training the classifier over dataset that discarded/pruned images with fewer than 100 images per class showed gains over a classifier that did not prune.

|  | Complete Train Set | Pruned Train Set |
|---|---|---|
| Val. Error | 0.88 | 0.86 |

Table 2: AlexNet validation error after 10 epochs

This verified that pruning classes which the classifier does not have a lot of information about helped the model to perform better during evaluation time.

All subsequent experiments utilize this form of data preprocessing.

5.2. Transfer Learning for weight initialization

We also perform a simple experiment to check the effectiveness of transfer learning to initialize weights, as well as to check if we should train all layers or only the last layer.

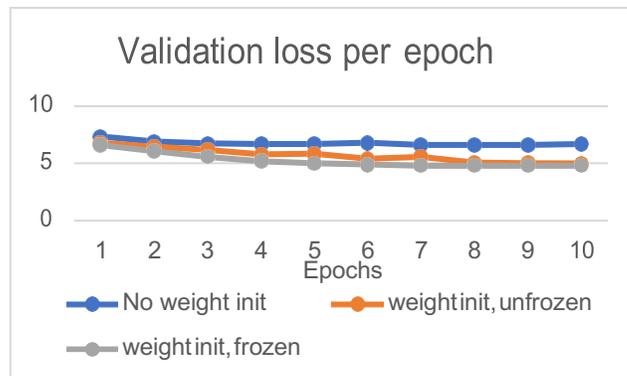

Figure 4: Validation loss using AlexNet

We see that validation loss for AlexNet with transfer learning and freezing all layers except the last performs best over 10 epochs.

All subsequent experiments used this form of transfer learning where we initialize weights by training on ImageNet 1k and freeze all layers except the last.

5.3 Using established CNN architectures

We experiment with the various architectures described above, trained over the pruned train set and using transfer learning, while freezing all layers except the last.

| Model Architecture | Validation Error |
|---|---|
| Alexnet | 0.86 |
| DenseNet | 0.71 |
| SqueezeNet | 0.80 |
| ShuffleNet | 0.96 |
| ResNext | 0.68 |

Table 3: Validation error after 10 epochs

Overall, ResNext has the least validation error of 0.68, followed by DenseNet coming a close second with 0.71. Shufflenet performs especially worse with a validation error of 0.96.

When evaluated on the test set, ResNext gave a similar 0.69 error rate, showing that the validation set and test set error were very close to each other and the training methods did not overfit.

6. Conclusions

We see that data pruning and transfer learning techniques work well when training models over the iNat 2019 dataset, likely meaning they would generalize well to other large scale classification tasks. Amongst the architectures we tested, ResNext works the best for large scale classification. It may have performed well due to its property to handle vanishing gradients well which could be a more prominent issue with large scale classification tasks. DenseNet also came in at a close second, possibly due to similar reasons.

7. Future Work

Since each of the architectures tested have different strengths, it would be interesting to see how these architectures perform in an ensemble framework which can capitalize on the strengths of each of the architectures. It would also be interesting to see how these models compare over an extended number of epochs exceeding 10.